\pdfoutput=1
\documentclass[11pt]{article}

\usepackage[final]{acl}

\usepackage{times}
\usepackage{latexsym}

\usepackage[T1]{fontenc}
\usepackage[utf8]{inputenc}

\usepackage{microtype}

\usepackage{inconsolata}

\usepackage{graphicx}

\usepackage{tabularx}
\usepackage{booktabs}

\title{\emph{Stop! In the Name of Flaws}: \\ Disentangling Personal Names and Sociodemographic Attributes in NLP}

\author{
 \textbf{Vagrant Gautam\textsuperscript{1}}\quad\quad
 \textbf{Arjun Subramonian\textsuperscript{2}}\quad\quad
 \textbf{Anne Lauscher\textsuperscript{3}}\quad\quad
 \textbf{Os Keyes\textsuperscript{4}}
\\
 \textsuperscript{1}Saarland University, Germany\quad
 \textsuperscript{2}University of California, Los Angeles, USA\quad
\\
 \textsuperscript{3}Universität Hamburg, Germany\quad
 \textsuperscript{4}University of Washington, USA
\\
}

\begin{document}
\maketitle
\begin{abstract}
Personal names simultaneously differentiate individuals and categorize them in ways that are important in a given society.
While the natural language processing community has thus associated personal names with sociodemographic characteristics in a variety of tasks, researchers have engaged to varying degrees with the established methodological problems in doing so.
To guide future work that uses names and sociodemographic characteristics, we provide an overview of relevant research:
first, we present an interdisciplinary background on names and naming.
We then survey the issues inherent to associating names with sociodemographic attributes, covering problems of validity (e.g., systematic error, construct validity), as well as ethical concerns (e.g., harms, differential impact, cultural insensitivity).
Finally, we provide guiding questions along with normative recommendations to avoid validity and ethical pitfalls when dealing with names and sociodemographic characteristics in natural language processing.\looseness=-1
\end{abstract}

\section{Introduction}
A person's identity is a complex and paradoxical thing - it simultaneously identifies someone's \textit{uniqueness}, and categorizes them, identifying what they have in common with others~\citep{strauss2017mirrors}. A perfect example of this phenomenon is a person's \textit{name}. Personal names are proper nouns used to refer to individuals. They play an important distinguishing role in our lives, as they let us uniquely represent people mentally, refer to them directly in speech, and underscore their significance as individuals~\citep{Jeshion_2009}. For these reasons, personal names are a linguistic universal, i.e., they appear across languages and cultures, although naming customs vary across the world~\citep{Hough_Izdebska_2016}.

% the connection of names to social variables
But alongside differentiating people,  names also categorize them in their society.
Names assigned to people often index aspects of identity that are important in the context of their society, including sex, religion, tribe, stage of life, etc.
Personal names are thus rich resources to understand the social organization of communities, and have been studied across anthropology \citep{alford1987naming,Hough_Izdebska_2016}, sociology \citep{Marx1999WhatsIA, Pilcher2017NamesA}, linguistics \citep{Allerton1987TheLA, anderson2003structure}, and onomastics \citep{AlvarezAltman1987NamesIL, Adams2009PowerPA}.

\begin{figure}[t]
    \centering
    \includegraphics[width=\columnwidth]{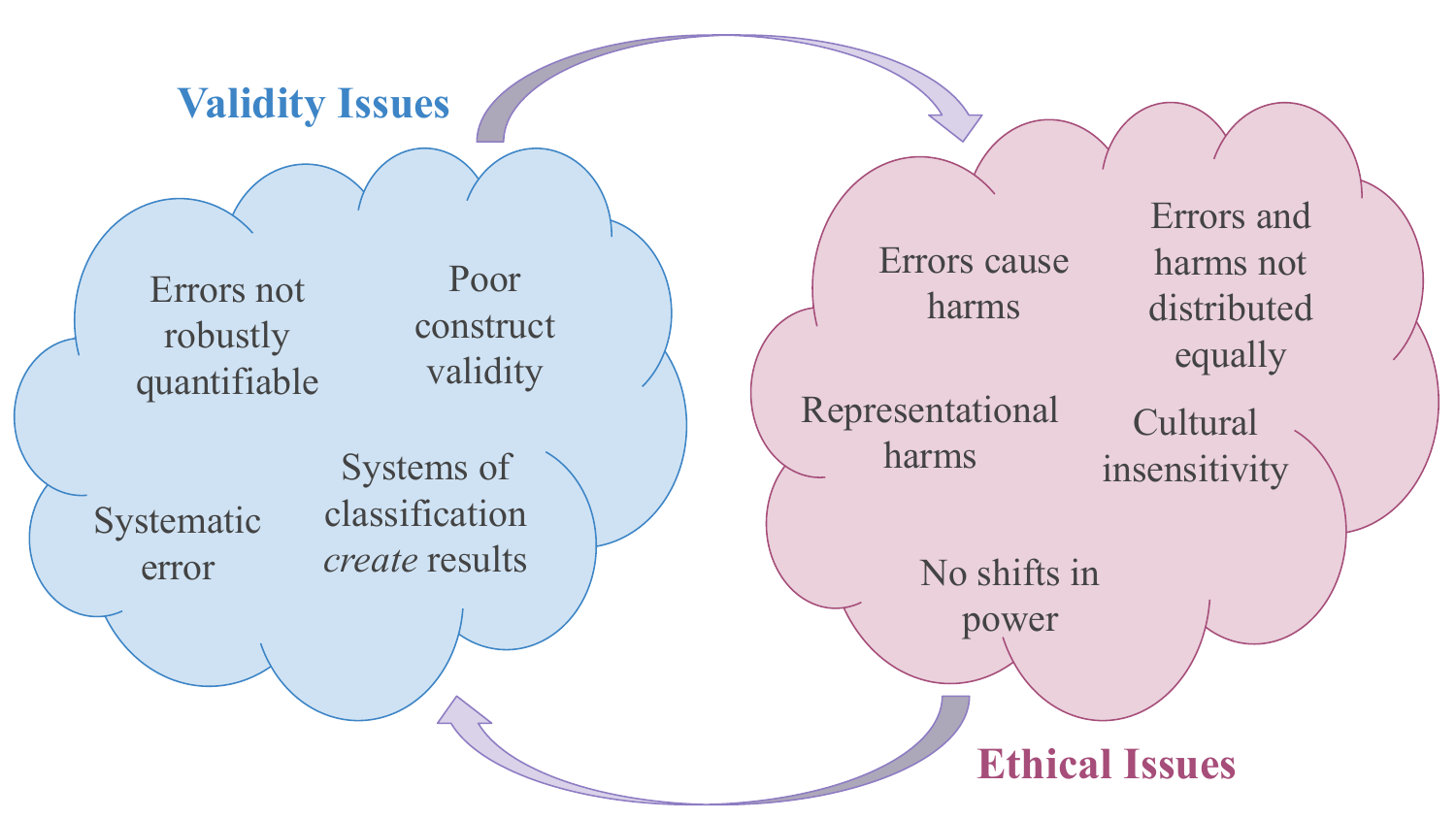} 
    \caption{Overview of the methodological issues (concerning validity and ethics) of the use of personal names and sociodemographic characteristics in NLP.}
    \vspace{-5mm}
    \label{fig:one}
\end{figure}

% what NLP does with this
In natural language processing (NLP) as well, personal names have a long history of use---NLP researchers have worked on identifying and disambiguating uses of personal names \citep{Mann2003UnsupervisedPN, minkov-etal-2005-extracting, Frber2022TheMA} and have examined name translation~\citep{sennrich-etal-2016-neural,wang-etal-2022-measuring,sandoval-etal-2023-rose} and name transliteration~\citep{li-etal-2007-semantic,benites-etal-2020-translit,saleva-lignos-2024-paranames-1}.
Increasingly, NLP researchers also use personal names along with sociodemographic characteristics for passive analysis of media and scholarly content~\citep{vogel-jurafsky-2012-said, knowles-etal-2016-demographer, mohammad-2020-gender, Asr2021TheGG}, or to examine model biases and harms~\citep{hall-maudslay-etal-2019-name, romanov-etal-2019-whats, webster2021measuring}.
However, these papers engage to varying degrees with concerns that have been raised outside of NLP about the methodological validity and ethics of associating names with sociodemographic characteristics.
We argue that neglect of these issues is a significant barrier to valid and respectful research, as well as more inclusive NLP systems.

Hence, we contribute an overview of the issues with associating names with sociodemographic attributes (focused on gender and race, two popular categories used in NLP research), as shown in Figure \ref{fig:one}.
We begin with background on names in other fields and in NLP (\S\ref{sec:background-names}), and lay out the problems with validity (\S\ref{sec:validity-problems}) and ethical concerns (\S\ref{sec:ethical-problems}) raised when associating personal names with sociodemographic characteristics. Finally, we present guiding questions along with normative recommendations (\S\ref{sec:recommendations}) to guide future work in the field with names.

\paragraph{Bias statement.}
We consider group-level, individual, and representational harms,
terms which we explain where we use them in Section \ref{sec:ethical-problems}.

\paragraph{Positionality statement.}

All authors have a background in data science and ethics, and one has a background in philosophy. Three of the authors are trans and have names that are likely unintelligible to popular name-based sociodemographic inference methods. Two authors are trans people of colour; as such, many examples in this paper reflect concerns about misgendering and racialization.

\section{Background}
\label{sec:background-names}
We begin with some background: on names and naming, and on the use of names in NLP.

\subsection{Names and Naming}

Names are generally regarded as social phenomena that serve two central functions that are sometimes in conflict: differentiation and categorization of individuals~\citep{alford1987naming}.
Differentiation is important psychologically and semantically for us to be able to directly refer to and mentally represent individuals, and names also serve to underscore their referent's significance as an individual~\citep{Jeshion_2009}.
Categorization, on the other hand, is important for the social organization of communities, and naming conventions tend to reflect factors that are important to a community at a given point in time, e.g., gender, religion, descent, transition to adulthood, and so on~\citep{Hough_Izdebska_2016}.
For instance, the practice of naming someone after their father or grandfather---patronymic naming---was once common across Europe, and was popular in Sweden until the nineteenth century (e.g., \textit{Samuelsson}) and continues into Iceland today (e.g., \textit{Gunnarsd\'{o}ttir})~\citep{Hough_Izdebska_2016}.
This example shows how names and naming can \textit{only} be understood in a specific (geographic, cultural, temporal) context, and even then includes a lot of variation.
As folk assumptions about names tend to overlook the wide variation in names and naming \citep{kalzumeusFalsehoodsProgrammers}, we present an overview of naming as it relates to sociodemographic characteristics below.

\paragraph{Variation in societal conventions.}
% Big-picture societal conventions vary a lot
The markers considered important to index in a name vary widely across cultures. For example, almost all European naming systems and indeed most societies across the world tend to assign sex-typed names~\citep{Hough_Izdebska_2016}, while South Indian naming conventions often index caste~\citep{Meganathan2009ThePO}.
However, convention does not mean that every single individual is assigned a name that neatly follows that convention, as shown by the long history of gender-ambiguous names in the  U.S.~\citep{barryharper1982evolution}.
Additionally, gendered associations for specific names change over time~\citep{Barry_Harper_1993}, as do naming conventions in societies--for example, it is becoming increasingly popular to assign non-gendered names in the  U.S. and in Israel~\citep{Hough_Izdebska_2016,Obasi_Mocarski_Holt_Hope_Woodruff_2019}.
Apart from conventions, names themselves are not static and unchanging from birth, with many names changing due to partnerships, adoption, transition to a different life stage or gender, and so on~\citep{Hough_Izdebska_2016,Obasi_Mocarski_Holt_Hope_Woodruff_2019,aiatsisIndigenousNames}.

\paragraph{Assimilation and resistance to convention.}
% Heterogeneous societies and power complicate these macro societal conventions among subgroups
% Factions of society subject to power relations do naming differently
Trends in big-picture naming conventions are complicated by factions of society who want to resist imposed classification.
Increasingly heterogeneous societies are a natural setting for such tensions;
cross-cultural associations with sociodemographic characteristics can differ and sometimes clash, complicating naming, e.g., names like \textit{Nicola} and \textit{Andrea} tend to be assigned to boys in Italy but to girls in Germany.
As Germany is a society with highly regulated naming practices, inclusion of these names necessitated a court judgment~\cite{Hough_Izdebska_2016}.
Immigrant families thus have to juggle the delicate balance of asserting their identity but avoiding name-based stigma and discrimination in the new culture.
Their naming practices have therefore been studied as an indicator of attitudes towards assimilation or its rejection, showing how names are not a transparent indicator of race~\citep{Sue_Telles_2007,becker2009immigrants}.
Even among adults, imperialism and colonialism are forces that affect naming.
Indigenous individuals have been forced to adopt Western names in settler colonial and postcolonial societies, e.g., the U.S., Canada, Australia~\citep{aiatsisIndigenousNames}. 
Similarly, Chinese individuals around the world adopt Western names in conversational \citep{LI1997489} and professional settings \citep{dwChineseSpeakers}.
Among trans and gender-nonconforming adults, many choose a new name to reflect and express their gender,
walking the tightrope between normativity and self-assertion~\citep{konnelly2021nuance};
\citet{Obasi_Mocarski_Holt_Hope_Woodruff_2019} find that 50\% of gender-nonconforming respondents who change their name pick a gender-neutral name.
Beyond transgender people, new names and pseudonyms are also often self-selected to assert agency in identity creation, e.g., bell hooks, Sojourner Truth, and Malcolm X~\citep{baker2021there}.\looseness=-1

\paragraph{Quantitative aspects of naming.}
As naming involves a trade-off between differentiation and categorization, names often recur, a quantitative assumption that a lot of sociological, anthropological and NLP classification relies on~\citep{alford1987naming}.
However, the distributions of names and people can be very different.
\citet{Weitman_1981} finds that in 100 years of first names from Israel's Population Registry, the most frequent names (101+ occurrences) account for the majority of the population of a society (91\%), but this corresponds to just a tiny minority of all assigned \textit{names} (2.93\%).
These numbers could vary widely depending on the society, as, for example, the Chuukese people of Micronesia have a tradition of giving entirely unique names to children~\citep{alford1987naming}.
Hence, it is important to distinguish when names are the object of study and when people are, to contextualize any results that involve the analysis of names.

\subsection{Names and Sociodemographic Characteristics in NLP}
Here, we present a non-comprehensive list of papers to illustrate some common uses of names and sociodemographic characteristics in NLP.

\paragraph{NLP tasks and problems.} Numerous NLP works have developed algorithms to infer sociodemographic attributes from names \citep{Chang2010ePluribusEO, Liu2013WhatsIA, knowles-etal-2016-demographer}, e.g., for passive analysis of social media content. Another line of NLP papers have relied on names to quantify gender disparities in academic publishing \citep{vogel-jurafsky-2012-said, mohammad-2020-gender} or media representation \citep{Asr2021TheGG}. Some NLP works have identified preserving dominant gender associations as an important criterion for transliteration and translation \citep{li-etal-2007-semantic, wang-etal-2022-measuring}.
Names are also used to investigate social biases in NLP systems and language models~\citep{kotek_et_al_2023,an-etal-2023-sodapop,ibaraki-etal-2024-analyzing-occupational}.
For example, \citet{dearteaga2019bios} study how first names, which they consider ``explicit gender indicators,'' affect the gender bias of occupation prediction from biographies.
Similarly, \citet{jeoung-etal-2023-examining} assess the causal impact of first names, which they posit ``may serve as
proxies for (intersectional) socio-demographic representations,'' on the commonsense reasoning performance of language models.
\citet{Smith2021HiMN} measure racial biases as well, evaluating generative dialogue models by having ``one conversational partner [\ldots] state a name commonly associated with a certain gender and/or race/ethnicity.''
In this line of research, it is commonplace to use skewed reference populations such as U.S. census data \citep{us-census} and Social Security Administration baby names \citep{ssa-baby-names} for gender assocations~\citep{Lockhart_King_Munsch_2023}.

\paragraph{Engagement with pitfalls.}
In these works, researchers engage to varying degrees with the established methodological and ethical problems of associating names with sociodemographic characteristics.
Some NLP papers make unfounded assumptions about names, e.g., \citet{vogel-jurafsky-2012-said} posit that certain names are ``unambiguous'' with respect to gender across languages, and \citet{wang-etal-2022-measuring} claim that there exist ``names with obvious gender.''
Other papers are more critically reflective, acknowledging the limitations of their work:
\citet{knowles-etal-2016-demographer} state that their classifier to predict gender from names is biased towards the U.S. and assumes gender is binary, but leaves these issues ``to be addressed in future work.''
\citet{mohammad-2020-gender} acknowledges that inferring gender from names can yield misgendering because ``names do not capture gender fluidity or contextual gender,'' but suggest a trade-off with ``the benefits of NLP techniques and social category detection.''
Encouragingly, some recent papers opt for more inclusive study designs after engaging deeply with the pitfalls of using names and sociodemographic characteristics~\citep{sandoval-etal-2023-rose,saunders-olsen-2023-gender,lassen-etal-2023-detecting}.

\section{Validity Issues}
\label{sec:validity-problems}

In this section, we present issues of validity when associating names with sociodemographic categories, or using names to infer them.
Issues of validity mean that results with these operationalizations may neither be indicative of what we actually want to measure, nor of reality.

\paragraph{Error is not quantifiable without asking humans.}
The accuracy of using names to infer sociodemographic characteristics of individuals cannot be quantified without ground truth data, which for people's identities, can \textit{only} be obtained by asking them.
Multiple studies thus empirically analyze the error rates of name-based gender and race inference systems as compared to gold data in different contexts~\citep{Karimi_Wagner_Lemmerich_Jadidi_Strohmaier_2016,kozlowski2022avoiding,Van_Buskirk_Clauset_Larremore_2023,Lockhart_King_Munsch_2023}.\footnote{All these studies look at imputing an individual's gender, but the gold labels they compare to are, confusingly, not always self-reported gender! Some use gender assigned by annotators as the ground truth, which would be fine if comparing to \textit{perceptions} of an individual based on their name, but these studies do not, raising further questions about their methodological validity.}
For example, \citet{Lockhart_King_Munsch_2023} evaluate gender and race inference systems using self-reported data from nearly 20,000 individuals.
Importantly, their self-reported data does not directly transfer to other contexts, as their respondents are authors of English language social science journal articles who are mostly located in the U.S.
Using this data as reference data for a system with users located primarily in India, or for U.S. authors in a different century, makes little sense.
In new environments, it is simply not possible to reasonably estimate the bounds of error of a name-based analysis, and results without a corresponding analysis of self-reported data should not be taken seriously.

\paragraph{Popular design choices lead to systematic error and selection bias.}
Names that are uninformative of a sociodemographic characteristic present an issue for tools that aim to label everyone.
In the context of gender, names like \textit{Alex} have no unique gendered association in the U.S. and Canada; with race, names assigned by Black and white parents overlap in the U.S.~\citep{Lockhart_King_Munsch_2023}, and religious names are used around the world~\citep{Curtis2005AfricanAmericanIR, 
Ikotun2014NewTI}); at the intersection of gender and race, many Chinese names are not gender-associated when Romanized, and infrequent names are also not informative.
Two common design choices for handling uninformative names are to assign the majority class label anyway, or, alternatively, to just exclude them.
Assigning the majority class (i.e., classifying \textit{all} people named \textit{Miaoran} as female if a gender prediction tool predicts the name to be ``60\% female'') results in systematic error~\citep{Kirkup_Frenkel_2006}.
On the other hand, excluding uninformative names from the analysis completely alters the makeup of the data and therefore the results~\citep{Mihaljević_Tullney_Santamaría_Steinfeldt_2019}, resulting in selection bias.
Both choices affect internal validity, i.e., gaps in the translation from measurements to overall conclusions~\citep{liao2021are}, leading to less robust and trustworthy results.

\paragraph{Poor construct validity.}
Construct validity asks how well an abstract concept can be measured through some indicator~\citep{messick1995standards};
in our case, the question is: how valid is it to assign sociodemographic categories via names?\footnote{While we focus on the construct validity of names in this section, we note that poor construct validity also applies to the sociodemographic categories themselves~\citep{benthall2019racialcategories,hanna2020towards} and to abstract concepts such as ``bias'' and ``fairness,'' which show up frequently in the study of names and sociodemographic categories in NLP~\citep{blodgett-etal-2020-language,Jacobs_Wallach_2021}.}
The answer to this depends on what aspects of the sociodemographic category we are interested in: identity, socialization, expression, perception---all of which could differ and are frequently conflated~\citep{keyes2021youkeep}.
As discussed previously, many names are simply not informative of certain sociodemographic \textit{identities} in given contexts and with homogeneous populations;
\citet{Lockhart_King_Munsch_2023} find that overall error rates of name-based gender and race imputation tools range from 4.6\% to 86\% overall, and up to 100\% for particular subgroups, depending on the tool.
However, when it comes to the \textit{perception} of names as indexing a sociodemographic category, some names may have stronger construct validity, an assumption used by \citet{sandoval-etal-2023-rose} in their examination of names assigned at birth that are strongly associated with the baby's sex and the parents' race/ethnicity.
On the other hand, \citet{mohammad-2020-gender} uses names to operationalize both identity (to investigate trends in authorship) and perception (to investigate trends in citation) in a bibliometric analysis of the ACL Anthology, even though these need not match, and many underrepresented names are uninformative of identity as well as perception~\citep{Van_Buskirk_Clauset_Larremore_2023}.
As names do not neatly line up with sociodemographic identities, perceptions, or experiences in a context-independent way, it is critical to investigate construct validity of names in any setting where they are used.

\paragraph{Systems of classification \textit{create} results.}
Although classification is inherently human, classification systems are produced by culture and politics and end up \textit{creating} a view of the world~\citep{bowkerstar2000sorting}.
In computing, researchers have power and our positionality shapes how we view and operationalize categories of classification such as race and gender~\citep{Scheuerman_Wade_Lustig_Brubaker_2020,Scheuerman_Brubaker_2024}.
However, many such categories are unstable and contested \citep{keyes2021youkeep, Mickel2024RacialEthnicCI}.
For instance, it has been shown that different ways of operationalizing race can result in entirely different conclusions~\citep{Steidl_Werum_2019,benthall2019racialcategories,hanna2020towards}.
Individuals and groups thus cannot be treated as monoliths that can be characterized one-dimensionally via names.

\section{Ethical Issues}
\label{sec:ethical-problems}

The issues we have examined so far impact the scientific validity of claims made using personal names and sociodemographic categories.
Many of these problems arise from assumptions that can also be criticized on ethical grounds, as we show.

\paragraph{Errors cause harms.}
Harms can be broadly described as a setback in the interests or progress of people due to, e.g., the outcomes of an automatic process~\citep{feinberg1984harmless}.
Group-level harms are experienced collectively by people in a sociodemographic group, while individual harms (which might result from group membership) are experienced at the person-person or person-technology level.
Inferring gender from names frequently misgenders trans people and erases non-binary people~\citep{keyes2019misgendering}.
This perpetrates group-level erasure, as well as individual harms including damaging autonomy and dignity~\citep{Mcnamarah2020Misgendering}, inflicting psychological harms~\citep{dev-etal-2021-harms}, and a failure to show recognition respect to people~\citep{Darwall1977TwoKO}.
Certain types of name-based classification (e.g., of persecuted ethnic or religious groups) can threaten individual safety, and when NLP infrastructure is used for surveillance and targeting, this also threatens the safety of entire groups of people~\citep{wadhawan2022let}.\footnote{Regulation efforts such as the AI Act~\citep{Commission_2021} in the EU try to mitigate this, but this does not apply to authoritarian regimes' use of such technology~\citep{Briglia_2021}.}
NLP systems reinforce group-level structural discrimination in other ways as well;
name-based studies of racial disparities in academia have been shown to systematically discount the intellectual contributions of Black researchers~\citep{kozlowski2022avoiding}.

\paragraph{Errors and harms are not distributed equally.}
In their work on name-based gender classification, \citet{Van_Buskirk_Clauset_Larremore_2023} note that for names with no available data, assigning the majority class (in their case, male) maximizes accuracy, but results in 0\% error for the male class and 100\% error for any other classes.
For non-binary people, who are generally excluded from gender classification by design, the error rate is also almost always 100\%.
As for name-based race/ethnicity classifiers, \citet{Lockhart_King_Munsch_2023} show that people who self-identify as Filipino, Black, or Middle Eastern and North African, are misrecognized 55-75\% of the time, as compared to those who identify as white, Chinese, or Korean, who are mislabelled less than 10\% of the time.
As described above, misrecognition errors cause harms, which are then disproportionately experienced by these individuals.
We echo the conclusions of \citet{Mihaljević_Tullney_Santamaría_Steinfeldt_2019} and \citet{Lockhart_King_Munsch_2023}, i.e., that inclusive analyses are only possible when names are no longer used as a proxy to infer individuals' gender or race/ethnicity.

\paragraph{Representational harms.}
The erasure of identities and the flattening of variation in naming customs leads to representational harms, which include the reinforcement of essentialist categories and power structures~\citep{chien2024beyond}.
These harms primarily affect sociodemographic groups, e.g., non-binary people, who are often incorrectly and unjustly treated as a novel social phenomenon.
Groups of people with a certain name are often subject to a different type of representational harm, i.e., stereotyping.
For instance, the name \textit{Kevin} is associated with lower socioeconomic class in Germany~\citep{kaiser2010kevin}.
This stereotype, if encoded in an NLP system, could lead to quality-of-service differentials, as class is a sociodemographic characteristic that correlates with lower NLP performance in other contexts~\citep{curry2024classist}.

\paragraph{Cultural insensitivity.}
Conceptualizations of names and sociodemographic characteristics in NLP are often Western-centric, with folk assumptions about what names look like and the application of  U.S. racial categories and naming preferences to areas outside the  U.S., where they are unintelligible~\citep{field-etal-2021-survey}.
Non-Western naming practices are only sometimes described in papers where there is a specific language of study that is not English, e.g., name tagging in Arabic~\citep{shaalan-raza-2007-person} and Uyghur~\citep{abudukelimu-etal-2018-error}.
Even within English, there is little recognition of, e.g., English common nouns used as names in China~\citep[\textit{Billboard}, \textit{Shooting}, \textit{Pray}, etc.;][]{dwChineseSpeakers},
names containing spelling variations~\citep{aiatsisIndigenousNames},
and names that overlap in different cultures but have different associations, e.g., \textit{Jan} in the  U.S. compared to \textit{Jan} in Germany.
Beyond names, even gender, race, and other sociodemographic categories of relevance are different across cultures.
Many cultures have definitions of gender that go beyond the binary. Enforcing binary gender can thus be seen as an example of what \citet{Lugones2016} calls the ``coloniality'' of gender, which also results in epistemic violence, i.e., 
inhibiting people from producing knowledge, or silencing and discrediting their knowledge~\citep{chilisa2019indigenous}.

\begin{table*}[t]
    \centering
    \begin{tabularx}{\linewidth}{lX}
        \toprule
        \textbf{Theme} & \textbf{Guiding questions} \\
        \midrule
        Names vs. people & What are you aiming to study--names? Or people, via their names? \\
         % \midrule
         & What aspects of names are you interested in? \\
         % \midrule
         & What aspects of people are you interested in? \\
         \midrule
        Context & What is your context? \\
         % \midrule
         & Is processing names with NLP systems necessary to answer your questions? \\
         \midrule
        Harms and power & What kinds of harms apply? How can you mitigate them? \\
         % \midrule
         & Are you describing or prescribing? \\
         % \midrule
         & How does your work reify/redistribute power? \\
         \midrule
         Refusal & Is it still worth it? \\
         \bottomrule
    \end{tabularx}
    \caption{Our list of guiding questions for the use of names and sociodemographic categories in NLP, grouped by theme. See paragraphs in Section \ref{sec:recommendations} for detailed recommendations.}
    \label{tab:guiding-questions}
    \vspace{-1mm}
\end{table*}

\paragraph{No shifts in power.}
Names are a site for enforcing institutional power, as seen in ``real name'' policies~\citep{haimson2016constructing},
the (non-consensual) permanence of names in data infrastructure including Google Scholar~\citep{scholarhasfailedus}, governmental name regulation~\citep{diaPressReleases}, and the ``collective delusion'' of legal names, at least in the  U.S.~\citep{baker2021there}.
Names are also regulated socially through norms and expectations, many of which end up baked into our NLP systems.
We exercise power as NLP researchers and practitioners via our assumptions, which may reify sociodemographic categories, codify (or dismantle) associations between names and these categories, and create infrastructure that harms people at scale through surveillance or mislabelling. \citet{knowles-etal-2016-demographer} open-sourced their name-based gender inference tool, and \citet{vogel-jurafsky-2012-said} published (binary) gender labels with names of authors of NLP papers, which continue to be used in research~\citep{mohammad-2020-gender,Van_Buskirk_Clauset_Larremore_2023}.
This data reflects folk assumptions about gender, i.e., that it is binary, immutable and in perfect correspondence with names~\cite{keyes2019misgendering,cao-daume-iii-2021-toward}.
These datasets also deadname and misgender scientists from the NLP community, some of whom have spoken about its harms~\citep{sabrina_comms}.
Transgender people can only be counted in such a system if they conform to normative expectations~\citep{johnson2016transnormativity,konnelly2021nuance}, and if not, the burden is disproportionately on them to seek redress.
Even \citet{Asr2021TheGG}---a system relying on name-based gender inference that considers gender beyond the binary and does not publicly misgender individuals---does not shift power, as workarounds are a patch rather than built-in to the method;
gender inference still relies on APIs that use binary gender, and mistakes (typically, famous non-binary people) are manually corrected.
As all these examples show how power remains centralized, we echo previous calls to reimagine and reconfigure power relations in service of user autonomy~\citep{keyes2019insurrection,blodgett-etal-2020-language,hanna2020against}.\looseness=-1

\section{Guiding Questions and Recommendations}
\label{sec:recommendations}

In the previous sections, we have reviewed the myriad of issues surrounding the accuracy, validity and ethical use of names along with sociodemographic characteristics, and noted that all these issues arise from the same assumptions and inform each other.
In addition, we have shown that these problems apply overwhelmingly to those who are not cisgender, white, normatively named in a Western context, and well-represented in publicly available data.
Thus, work that uses names to operationalize people's sociodemographic categories most misrepresents and further marginalizes those who are already at the margins.
We take the normative position that this is not acceptable collateral damage, even (and especially!) in the name of ostensible fairness.
Thus, we come up with guiding questions and recommendations for NLP practitioners who are considering the use of names as they relate to sociodemographic categories. These are summarized in Table \ref{tab:guiding-questions}.

\paragraph{What are you aiming to study--names? Or people, via their names?}
It is acceptable to investigate what concepts NLP models associate with names, e.g., \textit{Madeleine} with \textit{kindness}. It is even acceptable to demonstrate that NLP models associate \textit{Marius} with the pronoun \textit{he} or with being male, and that these associations mirror common human associations \citep{Caliskan_Bryson_Narayanan_2017,Crabtree2023ValidatedNF}.
It is marginally acceptable to associate names with sociodemographic characteristics using imaginary people, e.g., drawing insights about gender bias more broadly based on how NLP models handle synthetic names of people assumed to be exclusively female;
while doing so does not compromise people's autonomy and dignity, it does further entrench hegemonic folk theories of names and people's identities, which has cultural harms.
Finally, it is unacceptable to present results about real people based solely on the assumption that their names provide a reliable signal about their identities, e.g., NLP papers authored by people named \textit{Madeleine} and \textit{Marius} cannot on their own provide trustworthy insights into gender and racial representation in the field, unless those specific individuals are asked about their gender.

\paragraph{What aspects of names are you interested in?}
Names are rich objects of study with variation in form, length, training data frequency, tokenization, associations, the strengths of these associations, and more.\footnote{Some of these aspects have already been explored in prior work in NLP~\citep{shwartz-etal-2020-grounded,wolfe-caliskan-2021-low,sandoval-etal-2023-rose}.}
Once you have decided what aspects to study, they must be operationalized and measured carefully, with attention to the context of the study or eventual system deployment.
This includes the scope of what counts as a ``name.''
For instance, considering the use of English common nouns as names (e.g., \textit{Cloud}) is particularly important when working with data from or systems deployed in China, where this naming practice is common~\citep{dwChineseSpeakers}.
Ensure that pre-processing choices are contextualized and do not distort results, that names are understood within context, and that error can be quantified robustly in the given context.
Thus, when measuring training data frequency of names, counting \textit{Cloud} tokens as names must consider when it is used as a name and when it is used simply as a noun.
Error could be quantified through manual analysis on a subset of the data.

\paragraph{What aspects of people are you interested in?}
People's identity and perceptions of them can differ, and these shape their experiences in various ways.
Therefore, it is first necessary to decide which aspects are relevant for a study.
\textit{Attempting to infer someone's identity using names is simply unacceptable due to the range of methodological and ethical concerns we list in this paper.}
We echo onomastic advice from nearly 40 years ago~\citep{Weitman_1981}, i.e., that ``inferences from names must be to the givers of these names, not to their bearers. What is more, inferences must always be to sociological formations (such as social classes, ethnic groups, historical generations, and the like), not to individual name-givers.''
In addition to studying formations of name-givers, it can also be acceptable to study perceptions of identity based on names.
For instance, numerous sociology papers have investigated racial and ethnic perceptions, as well as occupational stereotypes, based on names \citep{King2006WhatsIA, Gaddis2017HowBA, Gaddis2017RacialEthnicPF}.
Again, we emphasize that perceptions based on names are also highly contextual and non-universal.

\paragraph{What is your context?}
It is essential to understand the geographical, temporal, and cultural context of data with names, and document this information for datasets, e.g., with datasheets~\citep{gebru2021datasheets}.
What is the geographic, temporal, cultural and political context of the name data, name-bearers, models and sociodemographic categories you use?
Who are the people who will be impacted by your work, and what is their context?
What do you know about the naming practices in these contexts and the hetereogeneity in these practices?
Are you quantifying error with self-reported data?
We posit that it is unacceptable to use names without deeply engaging with context in these senses, and stress that ascribing contemporary Western identity categories to historical peoples without acknowledging the difference in contexts is reductive.

\paragraph{Is processing names with NLP systems necessary to answer your questions?}
For questions about human identity and perception based on names, NLP may not be the only or best method available.
We warn against technical solutionism~\citep{green2021contestation};
researchers should reflect on whether their questions could be approached with interviews, case studies, fourth-world paradigms, and so on~\citep{cameron2004evidence}.
Qualitative methods can provide deeper, richer evidence while respecting people's autonomy, dignity and context.
If your questions are instead about NLP systems, then processing names with them is certainly necessary, but we note that methodological pluralism and interdisciplinarity can enrich our practice as NLP researchers and practitioners regardless~\citep{wahle-etal-2023-cite}.\looseness=-1

\paragraph{What kinds of harms apply? How can you mitigate them?}
Our paper provides a starting point for harms that are relevant to the use of names and sociodemographic characteristics in NLP, and we encourage transparency about methodological and ethical problems~\citep{Bietti2019FromEW, technologyreview2020Lets}.
It is unacceptable to sideline these problems in the name of ``social good'' \citep{green2019good, greene2019better, Bennett2020Point}, and rather than treating entire segments of the world as limitations of or future work for your research, we encourage changing the methods themselves, as \citet{lauscher-etal-2022-welcome} do with neopronouns.
We recommend firmly grounding work in the ethical principles of autonomy, justice, and beneficence for people~\citep{Floridi2019Unified}, which we note are sadly under-represented in machine learning research~\citep{birhane-et-al-2022-values}.

\paragraph{Are you describing or prescribing?}
Descriptions of social phenomena are often conflated with normative behaviour (i.e., assumptions and assertions that create and reinforce norms) in NLP~\citep{vida-etal-2023-values}.
This is the subtle but significant difference between showing that sociodemographic name associations in language models mirror the judgements of some group of humans, versus stating that model associations \textit{should} mirror the judgements of some group of humans.
The latter ``cannot avoid creating and reinforcing norms''~\citep{talat-etal-2022-machine}.
Therefore, researchers should clearly distinguish descriptive and normative behaviours in the design, execution, and presentation of their experiments~\citep{vida-etal-2023-values}.
System designers do have to make decisions about how systems \textit{should} behave, i.e., they need to choose to perpetuate harmful structures in service of usability or to impose their own values on users and stakeholders when they take an advocacy position.
This is an ethical dilemma in design that participatory methods and feminist epistemologies are uniquely positioned to help with~\citep{Bardzell_2010}.

\paragraph{How does your work reify or redistribute power?}
Central to NLP and computer science at large are scale thinking~\citep{hanna2020against}, quantitative methodologies~\citep{birhane-et-al-2022-values}, and the illusion of objectivity~\citep{waseem2021disembodied}.
All these values serve to reify existing hierarchies and power structures.
We must first recognize our own power as NLP researchers and practitioners, and how our work can reinforce infrastructure for (mis)classifying real people and enable surveillance and harms at scale.
We recommend a counterpower stance~\citep{keyes2019insurrection}, situated knowledges~\citep{harraway}, and methods informed by a politic, e.g., intersectionality, a critical framework that centers justice, power, and reflexivity, and mandates praxis with teeth~\citep{collins2019intersectionality,Erete_Israni_Dillahunt_2018, ovalleetal2023intersectionality}.
Particularly for those of us who are interested in using NLP for social good, we should constantly be asking: ``social good for whom?''
The differential impact on people matters, and as researchers and practitioners, we have a responsibility to attend to it and resist the othering perpetuated by classification systems.

\paragraph{Is it still worth it?}
After considering all these guiding questions, we remind the reader that refusal is possible~\citep{never-again,what-wont-build,Lockhart_King_Munsch_2023,Mihaljević_Tullney_Santamaría_Steinfeldt_2019}, and indeed an important part of the history of science~\citep{genevaprotocol,united1978belmont,nuremberg-code}.\looseness=-1

\section{Related Work}

Several papers study and critically interrogate the inference and use of sociodemographic information in computing~\citep{larson-2017-gender,keyes2019misgendering,benthall2019racialcategories,hanna2020towards,keyes2021youkeep,field-etal-2021-survey,devinney2022theoriesofgender}, many of which touch upon names but do not address them in detail.
The work that deals with names in particular are all outside of NLP:
\citet{Karimi_Wagner_Lemmerich_Jadidi_Strohmaier_2016,keyes2017stop,Tzioumis2018DemographicAO,Mihaljević_Tullney_Santamaría_Steinfeldt_2019,scheuerman2019howcomputersseegender,Lockhart_King_Munsch_2023,Van_Buskirk_Clauset_Larremore_2023}.
These papers have different scopes and take a variety of positions with regards to the ethics of name-based inference, some of which we find insufficiently radical.
Finally, our recommendations echo those from prior work (particularly in the fields of human-computer interaction and science and technology studies), but are contextualized for names in NLP.
Among others, we take inspiration from \citet{keyes2019insurrection,hanna2020against,blodgett-etal-2020-language,scheuerman2020hci}; and \citet{green2021contestation}.

\section{Conclusion}
We present the field with an overview of names and naming as discussed in other disciplines.
We lay out background on naming practices around the world and describe how these practices create issues of validity (e.g., selection bias and construct validity) and ethical concerns (e.g., harms, cultural insensitivity), that affect NLP uses of names and sociodemographic characteristics.
Finally, we present a list of guiding questions and normative suggestions towards addressing these concerns in future work involving names in NLP.

% \clearpage

\section*{Acknowledgments}
We thank Lucy Li for recommending literature about personal names, sociodemographic characteristics, and social perceptions. We are also grateful to our reviewers, and to members of the Critical Media Lab Basel, Switzerland, and the Interdisciplinary Institute for Societal Computing, Germany, for their feedback on the ideas in this paper.

\section*{Limitations}
Our background on names and naming is limited, and meant only as a brief introduction to onomastics and related fields that use names and sociodemographic characteristics;
space prevents us from being more comprehensive and we refer the interested reader to our references for deeper discussion of onomastic variation.
Additionally, we know that problematic and decontextualized assumptions about names are rife within NLP based on our background as authors within or adjacent to the field, as well as writing in other fields about methods that are also popular in NLP.
However, as we do not undertake a comprehensive, critical survey of NLP papers that use names and sociodemographic characteristics, we cannot empirically quantify the extent to which the problems we outline plague NLP research, and we leave a more systematic study of this to future work.\looseness=-1

\bibliography{custom}

\appendix

\end{document}